\documentclass[conference]{IEEEtran}
\usepackage{times}

\usepackage[numbers]{natbib}
\usepackage{multicol}
\usepackage[bookmarks=true]{hyperref}
\usepackage{times}
\usepackage{graphics}
\usepackage{graphicx}
\usepackage{subfigure}
\usepackage{amsmath,amssymb,amsopn,amstext,amsfonts}
\usepackage[ruled,vlined]{algorithm2e}
\SetKwComment{Comment}{$\triangleright$\ }{}
\usepackage{cancel}
\usepackage{pdfsync}
\usepackage{balance}
\usepackage{color}
\usepackage{mathtools}
\usepackage{bm}

\usepackage{diagbox}
\usepackage{float}
\usepackage{epstopdf}
\usepackage{pifont}
\usepackage{fixltx2e}
\usepackage{amsmath}
\usepackage{multirow}
\usepackage{url}
\usepackage{verbatim}
\usepackage{caption}
\usepackage{adjustbox}
\usepackage{booktabs}
\usepackage{threeparttable}
\usepackage{makecell}
\usepackage{tabularx}
\usepackage[normalem]{ulem}

\newcommand{\etal}{\textit{et al}.}
\newcommand{\ie}{\textit{i}.\textit{e}.}
\newcommand{\eg}{\textit{e}.\textit{g}.}

\DeclareMathOperator*{\argmin}{arg\,min}

\global\long\def\bfa{\mathbf{a}}
\global\long\def\bfb{\mathbf{b}}
\global\long\def\bfc{\mathbf{c}}

\global\long\def\bfg{\mathbf{g}}

\global\long\def\bfn{\mathbf{n}}

\global\long\def\bfq{\mathbf{q}}
\global\long\def\bfr{\mathbf{r}}

\global\long\def\bft{\mathbf{t}}

\global\long\def\bfv{\mathbf{v}}

\global\long\def\bfx{\mathbf{x}}

\global\long\def\bfA{\mathbf{A}}
\global\long\def\bfB{\mathbf{B}}

\global\long\def\bfP{\mathbf{P}}

\global\long\def\bfR{\mathbf{R}}

\global\long\def\bfT{\mathbf{T}}

\global\long\def\ttA{\mathtt{A}}
\global\long\def\ttB{\mathtt{B}}

\global\long\def\ttM{\mathtt{M}}

\global\long\def\ttT{\mathtt{T}}

\global\long\def\ttW{\mathtt{W}}

\global\long\def\cC{\mathcal{C}}

\global\long\def\cS{\mathcal{S}}
\global\long\def\cT{\mathcal{T}}

\global\long\def\cX{\mathcal{X}}

\global\long\def\Rot{\mathtt{R}}

\global\long\def\cS{\mathcal{S}} %

\global\long\def\bfomega{\boldsymbol{\omega}}

\global\long\def\bfbeta{\boldsymbol{\beta}}

\global\long\def\bfgamma{\boldsymbol{\gamma}}
\global\long\def\bfOmega{\boldsymbol{\Omega}}

\newcommand{\revision}[1]{\textcolor{magenta}{#1}} %

\newcommand{\os}[5]
{
{}^{#1}_{#2}{#3}^{#4}_{#5}
}

\global\long\def\nflowx{\dot{\mathbf{x}}_n}
\global\long\def\mflowx{\dot{\mathbf{x}}}

\begin{document}

\title{Event-based Visual Inertial Velometer}

\author{\authorblockN{Xiuyuan Lu$^{1}$*,
Yi Zhou$^{2}$*,
Junkai Niu$^{2}$, 
Sheng Zhong$^{2}$, and
Shaojie Shen$^{1}$}
\authorblockA{$^{1}$CKS Robotic Institute, Hong Kong University of Science and Technology, Hong Kong, China\\}
\authorblockA{$^{2}$School of Robotics, Hunan University, Changsha, China\\}
\authorblockA{Email: xluaj@connect.ust.hk; eeyzhou@hnu.edu.cn (*equal contribution)}
}

\maketitle

\begin{abstract}
Neuromorphic event-based cameras are bio-inspired visual sensors with asynchronous pixels and extremely high temporal resolution.
Such favorable properties make them an excellent choice for solving state estimation tasks under aggressive ego motion.
However, failures of camera pose tracking are frequently witnessed in state-of-the-art event-based visual odometry systems when the local map cannot be updated in time.
One of the biggest roadblocks for this specific field is the absence of efficient and robust methods for data association without imposing any assumption on the environment.
This problem seems, however, unlikely to be addressed as in standard vision due to the motion-dependent observability of event data.
Therefore, we propose a map-free design for event-based visual-inertial state estimation in this paper.
Instead of estimating the position of the event camera, we find that recovering the instantaneous linear velocity is more consistent with the differential working principle of event cameras.
The proposed event-based visual-inertial velometer leverages a continuous-time formulation that incrementally fuses the heterogeneous measurements from a stereo event camera and an inertial measurement unit.
Experiments on both synthetic and real data demonstrate that the proposed method can recover instantaneous linear velocity in metric scale with low latency.
\end{abstract}

\IEEEpeerreviewmaketitle

\section*{Multimedia Material}
\noindent Supplemental video: {\revision{\small\url{{https://youtu.be/FCHFMGRj3So}}}\\

\section{Introduction}
\label{sec: introduction}

As opposed to standard cameras that capture synchronously the absolute brightness at a fixed frame rate, event cameras are bio-inspired sensors that acquire visual information in the form of a stream of asynchronous per-pixel intensity changes (called ``events'').
Endowed with micro-second temporal resolution, event cameras can sense at exactly the same rate as the scene dynamics (induced by either ego motion or independent moving objects).
Consequently, they do not suffer from motion blur and are qualified for state estimation tasks involving aggressive ego motion.
The main challenge of building such an event-based visual state estimator (a.k.a event-based visual odometry) is to solve in real time the tracking and mapping sub-problems by exploiting photometric and geometric constraints encoded in the generative model of events.

\begin{figure}[t]
    \centering
    \vspace{0.5cm}
    \includegraphics[width=0.95\linewidth]{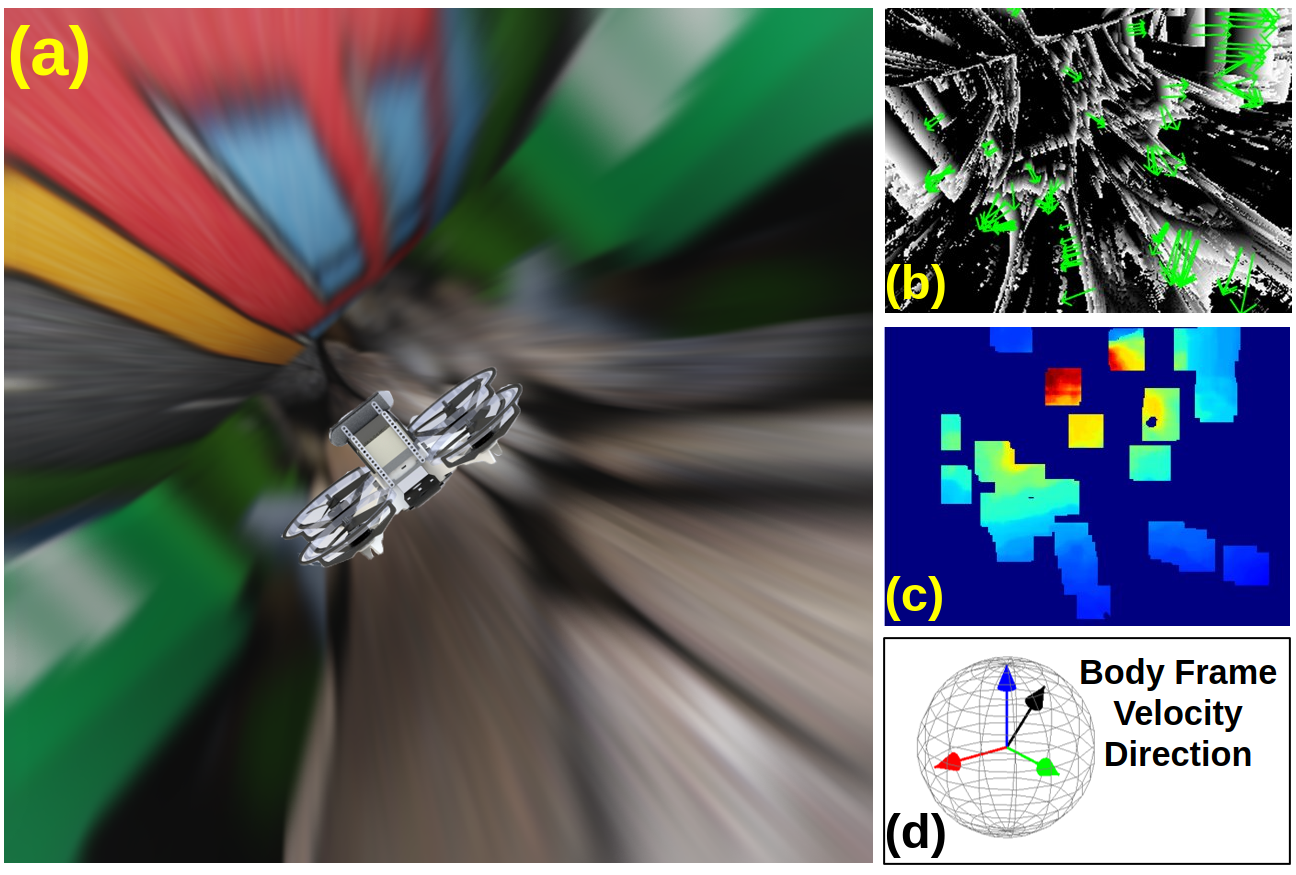}
    \caption{The proposed system takes as input the event data from a stereo event-based camera and inertial measurements from an IMU.
    (a) Aggressive maneuvers of a drone through a narrow corridor.
    (b) Event-based normal flow estimates.
    (c) Corresponding depth estimates.
    (d) Illustration of the normalized result of instantaneous linear velocity estimation.}
    \label{fig:eye_catcher}
\end{figure}

Although event-based methods for camera pose tracking \cite{gallego2017event, bryner2019event} have been proved effective in tackling aggressive ego-motion estimation, most event-based VO systems are not always qualified.
Early works using a monocular event-based camera (e.g.,~\cite{Kim16eccv, Rebecq17ral}) require a very gentle motion (typically a local-loopy behavior) for the initialization of a local 3D map, based on which the camera pose can be tracked using a 3D-2D registration pipeline.
To remove such a limitation on the initialization, Zhou \etal ~\cite{zhou2021esvo, niu2024imu} further use a stereo event-based camera to improve the efficiency and accuracy of mapping.
However, tracking failure is still witnessed when the ego motion of the camera suddenly becomes violent (mainly in terms of the angular velocity).
This is because the mapping sub-problem solver cannot update the local map in time.

To circumvent this issue, researchers resort to visual-inertial fusion and feature-based methods.
Although the event-inertial combination seems not as complimentary as the standard-visual-inertial counterpart, it is still beneficial to fuse event data with measurements of an inertial measurement unit (IMU).
First, an IMU's measurements can be used as a motion prior to propagate the estimated state.
Besides, such motion prior can be leveraged to obtain a motion-compensated image of warped events (IWE), on top of which a feature-based pipeline can be simply built \cite{Rebecq17bmvc}.
Obviously, the success of \cite{Rebecq17bmvc} largely relies on a sharp IWE, which is obtained via a geometric 3D-2D warping operation that requires the median depth of the local map.
This warping operation, however, can easily fail if there are relatively large depth variations.
To alleviate the front end's dependence on motion prior, some works \cite{hadviger2021feature,chen2023esvio} straightforwardly establish feature association on simple image-like representations of events (e.g., time surfaces \cite{Lagorce17pami}).
These ad hoc solutions may work to some extent, but the success of big-baseline feature matching is hardly guaranteed.
This is due to the fact that edges parallel to the epipolar line hardly trigger any events, and thus, junction-shape patterns may not be observed completely.
Consequently, the appearance similarity evaluated on such naive image-like representations could be violated in a sudden variation of linear velocity.

Noticing the dependence on the first-order kinematics, \cite{Manderscheid19cvpr, glover2021luvharris} propose a set of speed-invariant representations, which lead to constant thickness of edge patterns irrelevant to camera speed.
Though we have witnessed a lot trials on event-based feature detection and tracking~\cite{glover2021luvharris, Alzugaray18ral,Alzugaray18threedv,dardelet2021event}, most of them either cannot give satisfied feature tracking results as their standard-vision counterparts \cite{rublee2011orb, leutenegger2011brisk} do, or are computationally inefficient for real-time applications.
It's also worth mentioning that few works \cite{Chamorro20bmvc,chamorro2022event,Chamorro_2023_CVPR} manage to establish efficiently event-to-edge association, based on which parallel tracking and mapping is achieved under aggressive motion at an ultra-frame rate.
However, these solutions are only applicable in the presence of man-made geometric patterns.
All these issues drive us to think about a question: Is there a more rational design of state estimation for an event-based visual-inertial system that is more consistent with the differential working principle of event cameras?
To answer this question, we propose a first-order-kinematic state estimator, namely a velometer, using a stereo event-based camera and an IMU (see Fig.~\ref{fig:eye_catcher}).
Considering the special working principle of event cameras, we propose a continuous-time pipeline for the linear velocity estimation problem.
Specifically, our approach exploits the generative model of event-based normal flow induced by the first-order kinematics of the event camera.
Besides, the dynamic constraint from the IMU is also utilized.
The contribution of this paper is summarized as follows:
\begin{itemize}
\item A novel design of state estimator for an event-based visual-inertial system, which exploits the differential nature of event cameras and recovers the first-order kinematic state (i.e., linear velocity) by fusing events and inertial measurements.
\item A rigorous derivation for computing normal flow from spatial-temporal gradients of event data.
\item A continuous-time framework for event-based visual-inertial fusion, which can handle asynchronous event measurements and establish data association with temporally non-aligned measurements from the accelerometer.
\end{itemize}
The rest of the paper is organized as follows.
First, a literature review focusing on the topic of event-inertial state estimation is provided.
We then clarify the notion and revisit several key concepts as preliminaries in Sec.~\ref{sec:problem}.
Our method is discussed in Sec.~\ref{sec:method}, followed with the experimental evaluation in Sec.~\ref{sec:evaluation}.
Finally, the conclusion is drawn in Sec.~\ref{sec:conclusion}.

\section{Related Work}
\label{sec:related work}
Event-based state estimation intends to recover the motion parameters and a 3D map (optional) by exploiting specific constraints between ego motion and scene geometry in the generative model of event data.
Due to the special working principle and sensor characteristics, event cameras also appear as a complementary sensor in some visual-inertial systems aiming to address challenging scenarios that involve high-speed maneuvers and/or high-dynamic-range (HDR) illumination.
We also realize the existence of a large body of literature on the topic of event-based motion estimation, such as the optimization based methods for geometric model fitting on event data \cite{Gallego17ral, Gallego18cvpr, Nunes20eccv, zhou2021emsgc, nunes22pami, nunes2023time}, and learning methods that recover motion and structure \cite{zhu2018ev, zhu2019unsupervised}.
To be compact, our literature review focuses on two aspects: i) Fusion of event data and IMU measurements; and ii) Different choices of state variables in the problem of event-based state estimation.

\subsection{Event-based Visual-Inertial Odometry}
\label{subsec:event-based vio}
There is a large body of literature on this topic, most of which are built on top of a discrete-time back end that fuses events and IMU measurements in a manner of either probabilistic filtering or nonlinear optimization.
These methods differ mainly in the front tend, where event-based data association is established.
Early works \cite{Zhu17cvpr, Rebecq17bmvc} detect corners on an edge map generated by naively accumulating a set of events onto the image plane.
To keep tracking those corners, a two-step expectation-maximization (EM) method \cite{Zhu17cvpr} is proposed to estimate the event-based optical flow and update the feature template.
However, the computational complexity is so high that few successive reports apply this strategy in the front end.
With a prior knowledge of short-term relative motion and depth information of the local map, Rebecq \etal~\cite{Rebecq17bmvc} leverage the idea of motion compensation \cite{Gallego18cvpr} to restore sharp IWEs, on which corner features are constantly tracked.
More recently, an increasingly popular trend (e.g., \cite{hadviger2021feature, chen2023esvio}) is to apply traditional image-based feature detection and tracking methods on a certain image-like representation (e.g., \cite{Lagorce17pami, Manderscheid19cvpr}) of event data.
Despite their relative successes, the front end of these ad-hoc methods are not as theoretically sound as their standard-vision counterparts due to the motion-dependent nature of event data.
Moreover, the publicly available datasets \cite{Zhu18ral, gehrig2021dsec, gao2022vector, chaney2023m3ed} used for evaluation are typically collected using an agent undergoing gentle motion.
The superior performance of these methods lies in the improvement in terms of robustness under low-texture or HDR conditions.
Few of them (except for \cite{Rebecq17bmvc}), however, have ever demonstrated the ability to solve state estimation problems under aggressive maneuvers.

A relatively smaller body of literature looks into the design of the back end that can deal with asynchronous and temporally non-aligned data.
The employment of a continuous-time framework to the event-inertial fusion is firstly reported in \cite{Mueggler18tro}, which parametrizes the camera trajectories in SE(3) with cubic B-splines. 
Based on the generative model of event and inertial observations, the proposed probabilistic approach seeks the optimal estimate of the posteriori of the camera pose over a time interval, given the local map, event data, and IMU measurements.
The limitation of this method consists of, on one hand, the requisite of a known 3D map.
On the other hand, it does not manifest the ability to address aggressive ego motion estimation.
An alternative way to implement a continuous-time framework is using Gaussian process.
Wang \etal~\cite{wang2023event} represent a continuous-time trajectory with states defined at different timestamps of event feature tracklets.
By using a white-noise-on-acceleration motion prior, the smoothness of camera trajectories is enforced considering an agent's physical plausibility.
The overall trajectory is estimated using non-parametric Gaussian process regression.
Still, it does not manifest effectiveness under aggressive motion.
\subsection{First-Order Kinematics: A Better Choice?}
\label{subsec:choise of state variable}
It's only recently that researchers began to consider this question: What is the optimal choice for state variables in the problem of event-based state estimation?
Most existing event-inertial fusion pipelines target absolute camera poses.
However, this choice is inconsistent with the differential nature of event data.
To this end, Peng \etal~\cite{xin2021continuous} propose the first event-based state estimation method on the level of first-order kinematics.
The method relies on trifocal tensor geometry, which exploits the relationship between line features and first-order kinematics of the event camera.
With known angular velocity, up-to-scale linear velocity can be estimated in a closed form.
A successive work \cite{xu2023tight} is presented recently.
It extends \cite{xin2021continuous} by a direct solution of the linear velocity, and furthermore, delivers a tight event-inertial fusion back end achieving more reliable estimation of linear velocity.
These methods demonstrate their feasibility in tackling the problem of aggressive ego-motion estimation.
However, a limited scope of usage is clearly seen because of the dependence on straight lines, which are only available in man-made structural environments.
Besides, real-time performance is not demonstrated in \cite{xu2023tight}.

Considering the limitation of existing methods, we propose a novel state estimator at the level of first-order kinematics, namely a velometer, using a stereo event-based camera and an IMU.
Different from feature-based methods, our approach leverages a low-level observation, namely event-based normal flow, which is directly induced by the first-order kinematics of the event camera.
To be compatible with the asynchronous property of event data, we also employ a continuous-time framework using cubic B-splines to parametrize state variables.
The resulting pipeline is a map-free velocity estimator that tightly fuses event-based normal flow and IMU measurements, which can estimate linear velocity under aggressive motion in real time.
Furthermore, we demonstrate in Sec.~\ref{subsec: real-world experiments} that the proposed linear velocity estimator allows for recovering position information with only one integration operation, leading to more accurate dead-reckoning results.

\section{Problem Statement and Preliminaries}
\label{sec:problem}
Let $\os{}{}\cX{\ttA}{}$ denote a state $\cX$ being described in the coordinate system $\ttA$; the state $\cX$ can be position $\bfr$, linear velocity $\bfv$, and angular velocity $\bfomega$, etc.
The relative pose between two coordinate systems (e.g., the world frame $\ttW$ and the body frame $\ttB$) is denoted by a rigid transformation $\bfT_{\ttW\ttB} \doteq \{\bfR_{\ttW\ttB}, \bft_{\ttW\ttB}\}$, which consists of a rotation $\bfR_{\ttW\ttB}$ and a translation $\bft_{\ttW\ttB}$.
Given raw output from a stereo event camera and an IMU, the goal is to estimate real-scale linear velocity $\os{}{}\bfv{\ttB}{}$ of the sensor suite.
We assume that the stereo event-based camera has been pre-calibrated, including 
the spatial-temporal calibration between the left and right event cameras, and between the left camera and the IMU.
For simplification, we further assume that the left camera's coordinate system coincides with that of the IMU, which is denoted as the body coordinate system or body frame ($\ttB$).

To explain our method, we need to revisit several essential concepts as preliminaries, including \textit{motion flow}, \textit{normal flow}, and \textit{event-based depth estimation}.
\subsection{Motion Flow}
\label{subsec:motion flow}

\begin{figure}[t]
  \centering
  \includegraphics[width=\linewidth]{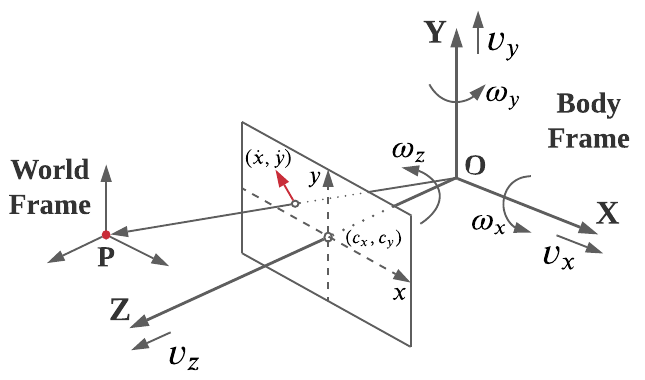}
  \caption{Geometry and kinematics involved in the problem of motion flow.
  An ideal perspective camera model is used to illustrate that the 3D point $\bfP$ is projected on the image plane.
  The components of both linear velocity and angular velocity are marked along their corresponding axis.
  The resulting motion flow is denoted by the red vector $(\dot{x},\dot{y})$.
  }
  \label{fig:geometry and kinematics}
\end{figure}
Motion flow, also known as optical flow, refers to derivatives of a 2D image location w.r.t time.
It describes how a 2D point would traverse on the image plane.
As illustrated in Fig.~\ref{fig:geometry and kinematics}, the motion flow of a textured 3D point $\os{}{}\bfP{\ttB}{} = [X,Y,Z]^{\ttT}$ can be determined according to the famous equation
\begin{align}
\label{eq:motion flow equation}
\begin{bmatrix}
    \dot{x}(t) \\
    \dot{y}(t)
    \end{bmatrix}
    &=
    \begin{bmatrix}
    \frac{v_{z}x'-fv_{x}}{Z} + \frac{\omega_{x}}{f}x'y' - \omega_{y}(f+\frac{{x'}^2}{f}) + \omega_{z}y' \\
    \frac{v_{z}y'-fv_{y}}{Z} + \omega_{x}(f+\frac{{y'}^2}{f}) - \frac{\omega_{y}}{f}x'y' - \omega_{z}x'
    \end{bmatrix}\nonumber \\
    &=
    \frac{1}{Z(\bfx)}\bfA(\bfx)\os{}{}{\bfv}{\ttB}{}(t) + \bfB(\bfx)\os{}{}{\bfomega}{\ttB}{}(t),
\end{align}
where $(\dot{x},\dot{y})$ denotes the motion flow vector at pixel $\bfx\doteq(x,y)$, 
$\os{}{}\bfomega{\ttB}{}=[\omega_{x},\omega_{y},\omega_{z}]$ the angular velocity,
$\os{}{}\bfv{\ttB}{}=[v_{x},v_{y},v_{z}]$ the linear velocity, and $f$ the focal length.
The relative image coordinates w.r.t the optical centre $(c_x,c_y)$ are denoted by $x'$ and $y'$, namely $x' \doteq x - c_{x}$ and $y' \doteq y - c_{y}$.

The motion flow equation (Eq.~\ref{eq:motion flow equation}) is first disclosed in~\cite{longuet1980interpretation}.
It is clearly demonstrated that the motion flow is uniquely determined by the 3D information ($Z$) and the first-order kinematics of the camera ($\os{}{}\bfv{\ttB}{}$ and $\os{}{}\bfomega{\ttB}{}$).
In other words, given the angular velocity $\os{}{}\bfomega{\ttB}{}$ (e.g., obtained from an IMU's measurements) at a time instant, the linear velocity $\os{}{}\bfv{\ttB}{}$ can be estimated if the motion flow and depth information at no fewer than two pixels are known.

\subsection{Normal Flow}
\label{subsec:normal flow}

For standard vision, the motion flow is typically determined according to the \textit{Horn-Schunck} model~\cite{horn1981determining}, which assumes that the brightness of a point remains constant when observed from two spatial and temporal neighbouring perspectives.
By expanding the image brightness function with first-order Taylor expansion, the brightness-constancy constraint goes as,
\begin{equation}
\label{eq:brightness consistency equation}
I(\bfx+d\bfx,t+dt) \approx I(\bfx,t) + \nabla_{\bfx} I(\bfx)
\begin{bmatrix}
dx\\
dy
\end{bmatrix}
+
\nabla_{t}I(\bfx) dt.
\end{equation}
Actually, this equation provides only one constraint such that only the partial component parallel to the image gradient direction can be determined as
\begin{equation}
\label{eq:normal flow equation 1}
    \dot{\bfx}_{n}(t) = -\frac{\nabla_{t}I(\bfx)}{\Vert \nabla_{\bfx} I(\bfx) \Vert^2} \nabla_{\bfx} I(\bfx),
\end{equation}
where $\dot{\bfx}_{n}$ denotes the projection of the flow $\dot{\bfx}$ on the image gradient direction $\nabla_{\bfx}I$.
This is referred to as the well-known \textit{aperture problem}, and the partial component $\dot{\bfx}_{n}$ is typically called \textit{normal flow}.

Let $\bfn$ denote the image gradient direction at $\bfx$, namely $\bfn = \frac{\nabla_{\bfx} I(\bfx)}{\Vert \nabla_{\bfx} I(\bfx) \Vert}$.
By multiplying both sides of Eq.~\ref{eq:motion flow equation} with $\bfn^{\ttT}$, we obtain
\begin{equation}
\label{eq:normal flow equation 2}
    \Vert \dot{\bfx}_{n}(t) \Vert
    = \frac{1}{Z(\bfx)}\bfn^{\ttT}\bfA(\bfx)\os{\ttB}{}{\bfv(t)}{}{} + \bfn^{\ttT}\bfB(\bfx)\os{\ttB}{}{\bfomega(t)}{}{}.
\end{equation}
Since normal flows rather than full motion flows can be estimated straightforwardly from raw events, Eq.~\ref{eq:normal flow equation 2} is used as a constraint in the estimation of linear velocity.
\subsection{Event-based Depth Estimation}
\label{subsec:event-based depth estimation}
Existing methods of recovering depth information from event data can be classified into two categories.
The first category is called the ``temporal stereo'' method, in which a monocular event-based camera is more often used.
Temporal stereo methods assume the camera's motion to be known as a prior \cite{Rebecq16bmvc, Rebecq18ijcv}.
They utilize events occurred over a temporal window to determine the 3D location of structures by searching the maximum in the disparity space image (DSI) through either voting in the discrete space \cite{Rebecq16bmvc, Rebecq18ijcv} or in a way of continuous optimization \cite{Gallego18cvpr}.
The second category is called the ``instantaneous stereo'' method, which typically applies a stereo event camera that have been calibrated extrinsically and synchronized temporally \cite{Kogler11isvc, Rogister12tnnls, CamunasMesa14fns}.
Typically, this kind of methods consists of two steps:
1) Searching corresponding events occurred in the left and right event cameras;
2) Triangulation.
To enhance the ratio of true-positive matching, state-of-the-art methods \cite{Ieng18fnins, Zhou18eccv} utilize a hybrid metric to measure the similarity of two events, which jointly considers temporal coherence, epipolar constraint, and motion consistency in the spatio-temporal neighborhood.
Since depth estimation is not the focus of this work, we simply apply the instantaneous stereo matching method (i.e.,~block matching on time surfaces \cite{Lagorce17pami}) implemented in \cite{zhou2021esvo}.
\section{Methodology}
\label{sec:method}
\begin{figure}[t]
  \centering
  \subfigure[t][\small{Temporal flow estimation as spatio-temporal planar fitting.}]
  {
  \includegraphics[width=0.45\columnwidth]{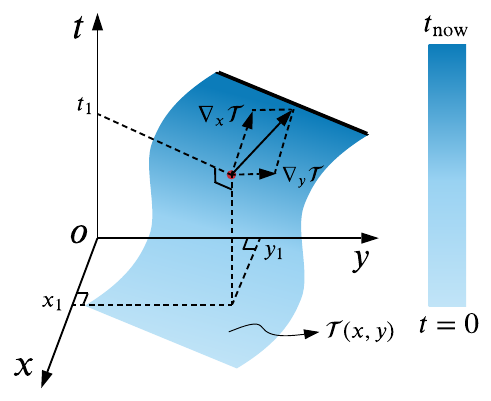}
  \label{subfig:temporal flow}
  }
  \subfigure[t][\small{Trigonometry in regard to motion flow $\dot{\bfx}$}, normal flow $\dot{\bfx}_n$ and brightness gradient direction $\bfn$.]
  {
  \includegraphics[width=0.45\columnwidth]{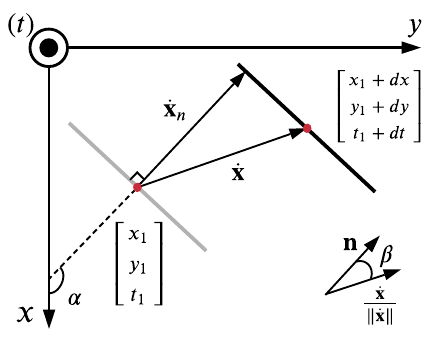}
  \label{subfig:normal flow}
  }
  \caption{General principles of normal flow estimation from raw event streams.
  (a) A time surface $\cT(x,y)$ is spanned in the spatio-temporal domain as the edge (black) traverses.
  The temporal flow $\nabla_{\bfx}\cT$ can be estimated by fitting a local plane.
  A color scale for the temporal values is provided aside. 
  The darker the color, the closer the time is to the current moment.
  (b) The normal flow $\dot{\bfx}_n$, namely the component of the motion flow $\dot{\bfx}$ along the direction of brightness gradient $\bfn$, is parallel to the temporal flow $\nabla_{\bfx}\cT$.}
  \label{fig:normal flow estimation from event}
\end{figure}

Given as input the raw events from a stereo event camera and the measurements from an IMU, the goal is to recover the instantaneous linear velocity in metric scale.
In this section, we first introduce our method of event-based normal flow estimation (\ref{subsec:event based normal flow estimation}).
Second, we present a continuous-time approach that fuses the two heterogeneous measurements by incrementally fitting a cubic B-spline in the space of linear velocity (\ref{subsec:continuous-time linear velocity estimation}).
Finally, we discuss our initialization method (\ref{subsec:initialization}).

\subsection{Event-based Normal Flow Estimation}
\label{subsec:event based normal flow estimation}
To estimate visual flows from raw events, \cite{Benosman14tnnls} proposes a local differential approach that fits local spatio-temporal planes on the time surface composed of co-active events.
The time surface is defined specifically by the function $\cT(x,y)$, which returns the timestamp by which the most recent event occurred at pixel $[x,y]^{\ttT}$.
The spatio-temporal plane's gradient is calculated as the temporal derivative w.r.t the pixel coordinate, namely $\nabla_{\bfx}\cT \doteq \begin{bmatrix}\nabla_x\cT,\nabla_y\cT\end{bmatrix}^{\ttT}$,
which reports the timestamp difference between spatially adjacent events in the direction of $x$ and $y$, respectively.

Let's assume that the brightness gradient direction is orthogonal to edges.
As shown in Fig.~\ref{subfig:temporal flow} and Fig.~\ref{subfig:normal flow}, the time surface $\cT$ can be regarded as a temporal profile induced by the normal flow $\nflowx$, and we have
\begin{equation}
\label{eq:relationship between temporal gradient and normal flow}
    \begin{bmatrix}
    \nabla_x\cT \\ 
    \nabla_y\cT
    \end{bmatrix}
    = \frac{1}{\Vert \dot{\bfx}_n \Vert}
    \begin{bmatrix}
    \cos{(\alpha)}\\
    \sin{(\alpha)}
    \end{bmatrix}.
\end{equation}
In fact, only normal flows could be recovered from temporal derivatives due to the aperture problem ($\beta$ is unknown in Fig.~\ref{subfig:normal flow}), and thus, the visual flow estimated by \cite{Benosman14tnnls} does not refer to the full motion flow $\mflowx$.
The method in \cite{Benosman14tnnls} is partially correct in the sense that the temporal gradient reveals the normal flow's direction.
The reciprocal of the temporal gradient's components would not straightforwardly give correct normal flows.
The amplitude of the normal flow can be determined using Eq.~\ref{eq:relationship between temporal gradient and normal flow}, and the normal flow vector is calculated as
\begin{equation}
\label{eq:normal flow computtion from events}
    \dot{\bfx}_n = \Vert \dot{\bfx}_n \Vert \bfn = \frac{1}{\sqrt{\nabla_x \cT^2 + \nabla_y \cT^2}}\bfn,
\end{equation}
where $\bfn = \frac{\nabla_{\bfx}\cT}{\Vert \nabla_{\bfx}\cT \Vert}$ refers to the direction of the normal flow\footnote{Note that we abuse the notation by duplicating $\bfn$, because the direction of the normal flow is parallel to that of the local image gradient.}.
We will justify our derivation by comparing against the result of~\cite{Benosman14tnnls} in the experiment (Sec.~\ref{subsec:evaluation result} and \ref{subsec: real-world experiments}).
It is noted that Valeiras \etal~\cite{valeiras2018event} have also pointed out the erroneous way of normal flow computation in \cite{Benosman14tnnls} and provided the correct formula.

\subsection{Continuous-Time Linear Velocity Estimation}
\label{subsec:continuous-time linear velocity estimation}
Although the computation of event-based normal flows cannot achieve the frequency as high as the streaming rate of event data (\eg~several million per second), it still can be much higher than the frame rate of a standard camera.
Besides, the computation can be carried out in an asynchronous manner, and thus, the resulting event-based normal flows typically do not coincide with the inertial measurements in time (\ie, temporally non-aligned data association).
Therefore, we employ a continuous-time formulation, which could bound the size of the optimization problem while enabling data association at any given time.
Specifically, we use a cubic B-spline based parametric model to represent the continuous-time linear velocity as
\begin{equation}
\label{eq:b-spline formula of velocity}
    \os{}{}{\bfv}{\ttB}{}(u(t)) = \sum_{i=0}^{3}B_{i,3}(u(t)) \os{}{}{\bfc}{}{i},
\end{equation}
where $B_{i,k}$ denotes the basis function, $i$ the index of control points, $k$ (=3) the order of the spline and $\os{}{}{\bfc}{}{i} \in \mathbb{R}^3$ the corresponding control point defined in the space of linear velocity.
The function $u(\cdot)$ is a normalization operator, which transfers time $t$ to the spline's parameter domain by means of basis translation~\cite{de1972calculating}.

Our goal is to fit a cubic B-spline that simultaneously satisfies the following two criteria: 1) The predicted normal flows are maximally consistent with the asynchronous normal flow measurements; and 2) The first-order derivative of the spline complies maximally with the acceleration measurements.
The first criterion can be simply evaluated using Eq.~\ref{eq:normal flow equation 2}.
The second criterion is evaluated, from another equivalent perspective, by calculating the difference between the predicted velocity increment and the pre-integration of acceleration in a local frame.

Let's denote the raw accelerometer and gyroscope measurements, $\tilde{\bfa}_t$ and $\tilde{\bfomega}_t$, in the body frame at time $t$ by
\begin{equation}
\label{eq:raw IMU measurement}
\begin{aligned}
    \tilde{\bfa}_t &= \bfa_t + \bfb_{a_t} + \Rot_{\ttW}^{t}\bfg^{\ttW} + \boldsymbol{\sigma}_a,\\
    \tilde{\bfomega}_t &= \bfomega_t + \bfb_{\omega_t} + \boldsymbol{\sigma}_w,
\end{aligned}
\end{equation}
where $\bfb_a$ and $\bfb_{\omega}$ are the accelerometer and gyroscope biases, while $\boldsymbol{\sigma}_a$ and $\boldsymbol{\sigma}_w$  the corresponding additive noise.
Let's further consider the IMU pre-integration~\cite{Forster15rss} during the time interval $[t_i, t_{i+1}]$.
Assuming the IMU biases are known, we integrate the inertial measurements in local frame $\ttB_i$ (the body frame at time $t_i$) as
\begin{equation}
\label{eq: imu pre-integration expression}
\begin{aligned}
    \tilde{\bfbeta}^{\ttB_i}_{\ttB_{i+1}} &= \int_{t_{i}}^{t_{i+1}}\Rot^{\ttB_i}_t(\tilde{\bfa}_t - \bfb_{a_t} - \bfn_a)dt, \\
    \tilde{\bfgamma}^{\ttB_i}_{\ttB_{i+1}} &= \int_{t_{i}}^{t_{i+1}}\frac{1}{2}\bfOmega(\tilde{\bfomega}_t - \bfb_{\omega_t} - \bfn_w) \bfgamma^{\ttB_i}_t dt,
\end{aligned}
\end{equation}
where $\bfOmega (\bfomega) = \begin{bmatrix}
-\left[\bfomega \right]_{\times} & \bfomega \\
-\bfomega^\ttT & 0
\end{bmatrix}$. 
Then the velocity and orientation can be propagated from $t_i$ to $t_{i+1}$ in the world frame by
\begin{equation}
\label{eq: imu pre-integration propogation}
\begin{aligned}
   \bfv^\ttW_{\ttB_{i+1}} &= \bfv^\ttW_{\ttB_{i}} - \bfg^\ttW\Delta t_i + \Rot^{\ttB_i}_{\ttW} \bfbeta^{\ttB_i}_{\ttB_{i+1}}, \\
   \bfq^{\ttW}_{\ttB_{i+1}} &= \bfq^{\ttW}_{\ttB_i} \otimes  \bfgamma^{\ttB_i}_{\ttB_{i+1}},
\end{aligned}
\end{equation}
where $\bfq$ refers to the orientation in the form of quaternion.
Consequently, the predicted velocity increment during the time interval can be calculated as
\begin{equation}
\label{eq:prediction of velocity increment}
    \hat{\bfbeta}^{t_i}_{t_{i+1}} = \Rot(\tilde{\bfgamma}^{t_i}_{t_{i+1}})\os{}{}{\bfv}{\ttB_{i+1}}{\ttB_{i+1}} + \os{}{}\bfg{\ttB_i}{}\Delta t_i - \os{}{}{\bfv}{\ttB_i}{\ttB_i}.
\end{equation}

Now let's discuss the formulation of the continuous-time state estimation problem.
The full state vector is defined as:
\begin{equation}
\label{eq: state vector}
\begin{aligned}
    \bf{\cX} &= [\bfc_0, \bfc_1, \cdots \bfc_n, \bfb_{0}, \bfb_{1}, \cdots \bfb_{n-3}], \\
    \bfb_{k} &= [\bfb_{a_k}, \bfb_{\omega_k}], k \in [0, n-3],
\end{aligned}
\end{equation}
where $n$ denotes the total number of control points, and $\bfb_{k}$ the IMU bias in the $k_{\tt{th}}$ B-spline segment.
Note that there are $n - 3$ cubic B-spline segments in the optimization.
Since each segment of the B-spline is short, we simply assume the biases are constant within each segment.

Combining Eq.~\ref{eq:normal flow equation 2}, Eq.~\ref{eq: imu pre-integration expression} and Eq.~\ref{eq:prediction of velocity increment}, we finally create the following objective function,
\begin{equation}
\label{eq: objective function}
\begin{aligned}
\bf{\cX}^{\star} &= \argmin_{\bf{\cX}} \cC,
\end{aligned}
\end{equation}
where the cost is defined as
\begin{equation}
\label{eq:new energy function}
\begin{aligned}
    \cC &\doteq 
    \sum_{t \in \cS} 
    \Vert \Vert \dot{\bfx}_{n}(t) \Vert - \frac{\bfn^{\ttT}\bfA(\bfx)\os{}{}{\bfv}{\ttB}{}(t)}{Z_t(\bfx)} - \bfn^{\ttT}\bfB(\bfx)\os{}{}{\bfomega}{\ttB}{}(t)\Vert^2_{\bfP_c}\\ 
    &+ 
    \sum_{i}^{M} \Vert \tilde{\bfbeta}^{t_i}_{t_{i+1}} - \Rot(\tilde{\bfgamma}^{t_i}_{t_{i+1}})\os{}{}{\bfv}{\ttB_{i+1}}{\ttB_{i+1}} - \os{}{}{\hat{\bfg}}{\ttB_i}{}\Delta t_i + \os{}{}{\bfv}{\ttB_i}{\ttB_i} \Vert^2_{\bfP^i_{i+1}}. 
\end{aligned}
\end{equation}
These two terms implement the above-mentioned two criteria, where $\cS$ denotes the timestamp set of all involved normal flow estimates, and $M$ the number of time intervals for IMU pre-integration.
To obtain a maximum a posteriori estimate, we apply the Mahalanobis norm to both measurement residuals.
The normal-flow covariance $\bfP_c$ is set empirically as a fixed one.
The covariance of the IMU measurements $\bfP^i_{i+1}$ is derived from the dynamics of error terms of Eq.~\ref{eq: imu pre-integration expression}~\cite{Qin18tro}.
Note that the orientation is not included in state vector $\bf{\cX}$.
Thus, we simply propagate the orientation by integrating raw angular velocities, given the initial orientation.
The gravity in the body frame ($\os{}{}{\hat{\bfg}}{\ttB_i}{}$) can be constantly estimated using the body-frame orientation via
$\os{}{}{\hat{\bfg}}{\ttB_i}{} = \Rot^{\ttB_i}_{\ttW} \os{}{}{\bfg}{\ttW}{}$.

\subsection{Initialization}
\label{subsec:initialization}
To initialize the non-linear optimization problem (Eq.~\ref{eq:new energy function}), we propose a linear way to coarsely determine the instantaneous linear velocity. 
Given as input the calculated normal flows\footnote{Note that the normal flows used here are those calculated within a short time interval. In other words, their timestamps can be deemed identical.}, their corresponding depth information, and the corresponding measurements of instantaneous angular velocity, a linear system with as unknown linear velocity $\os{\ttB}{}\bfv{}{}$ can be established based on Eq.~\ref{eq:normal flow equation 2}:
\begin{equation}
\label{eq:linear system for solving linear velocity}
\begin{bmatrix}
\bfn_{1}^\ttT \bfA_1\\
\vdots\\
\bfn_{K}^\ttT \bfA_K
\end{bmatrix}
\os{\ttB}{}\bfv{}{}
=
\begin{bmatrix}
Z_1(\Vert \dot{\bfx}_{n,1} \Vert - \bfn_1^{\ttT}\bfB_1\os{\ttB}{}\bfomega{}{})\\
\vdots\\
Z_K(\Vert \dot{\bfx}_{n,K} \Vert - \bfn_K^{\ttT}\bfB_K\os{\ttB}{}\bfomega{}{})
\end{bmatrix}.
\end{equation}
A minimal solver of Eq.~\ref{eq:linear system for solving linear velocity} requires three observations, and we use RANSAC~\cite{Fischler81cacm} for robust estimation.

\section{Experiments}
\label{sec:evaluation}

In this section, we first disclose the implementation details of our pipeline (Sec.~\ref{subsec:implementation details}).
Second, we introduce the datasets used in the experiments (Sec.~\ref{subsec: datasets}), followed with both qualitative and quantitative evaluation results and comparison against alternative solutions (Sec.~\ref{subsec:evaluation result} and Sec.~\ref{subsec: real-world experiments}).
Finally, we discuss the computational performance (Sec.~\ref{subsec: computational performance}) and discuss the limitation of the work (Sec.~\ref{subsec:discussion}).
\subsection{Implementation Details}
\label{subsec:implementation details}

\begin{figure}[t]
    \centering
    \includegraphics[width=1\linewidth]{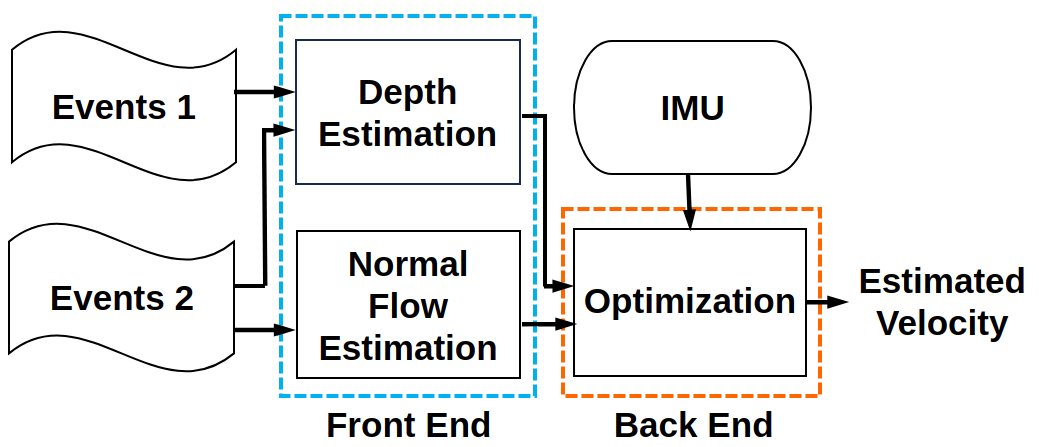}
    \caption{Flowchart of the proposed event-based visual-inertial velometer system.
    The system takes as input the events from a stereo event camera and an IMU's inertial measurements, and reports the estimated linear velocity.
    }
    \label{fig:pipeline}
\vspace{-0.5cm}
\end{figure}

The proposed event-based visual-inertial velometer system consists of two parts, as shown in Fig.~\ref{fig:pipeline}.
The front end implements the computation of event-based normal flow and corresponding depth.
Particularly, the event-based normal flow is computed according to Eq.~\ref{eq:normal flow computtion from events}, given a batch of events as input. 
To achieve real-time performance, the spatio-temporal plane fitting operation is conducted only on a small portion of events, while the rest are culled according to the following rules:
\begin{itemize}
\item Events whose pixel coordinates are close (smaller than 5 pixel) to the image plane's boundary are discarded;
\item Events with no more than 15 neighbouring data points in the local spatio-temporal volume defined by a $5 \times 5$ patch centred at the event's pixel coordinate are discarded;
\item Events whose timestamp differs from the average timestamp of the neighbouring events by more than a threshold (5\% of the event batch's duration) are discarded.
\end{itemize}
In this way, only 1\% - 2\% of the events are selected for subsequent normal flow estimation. 
We observe that obtaining 500 normal flow results approximately takes 5 ms.
Note that our implementation supports controlling the size of the event batch in two ways: 1) Using a constant number of events to adapt to the event streaming rate; 2) Using a constant-length time window to ensure high-frequency execution. 
In our implementation, the normal flow computation is triggered by the occurrence of every 45k events. 
The corresponding depth information is computed as mentioned in Sec. 
III-C with a maximum disparity of $48$ pixels and a block size of $17 \times 17$ pixels.
The back end implements the continuous-time nonlinear optimization pipeline using a cubic B-spline.
Specifically, a uniform rational B-spline is employed, and we set the knot interval to $0.1$ s.
The IMU pre-integration interval is set to $0.03$ s.
Ceres Solver~\cite{ceres-solver} is used to solve the non-linear optimization problem (Eq.~\ref{eq: objective function}).
\subsection{Datasets}
\label{subsec: datasets}
\begin{figure}[t]
    \centering
    \subfigure[Circular corridor.]{\includegraphics[width=0.47\columnwidth]{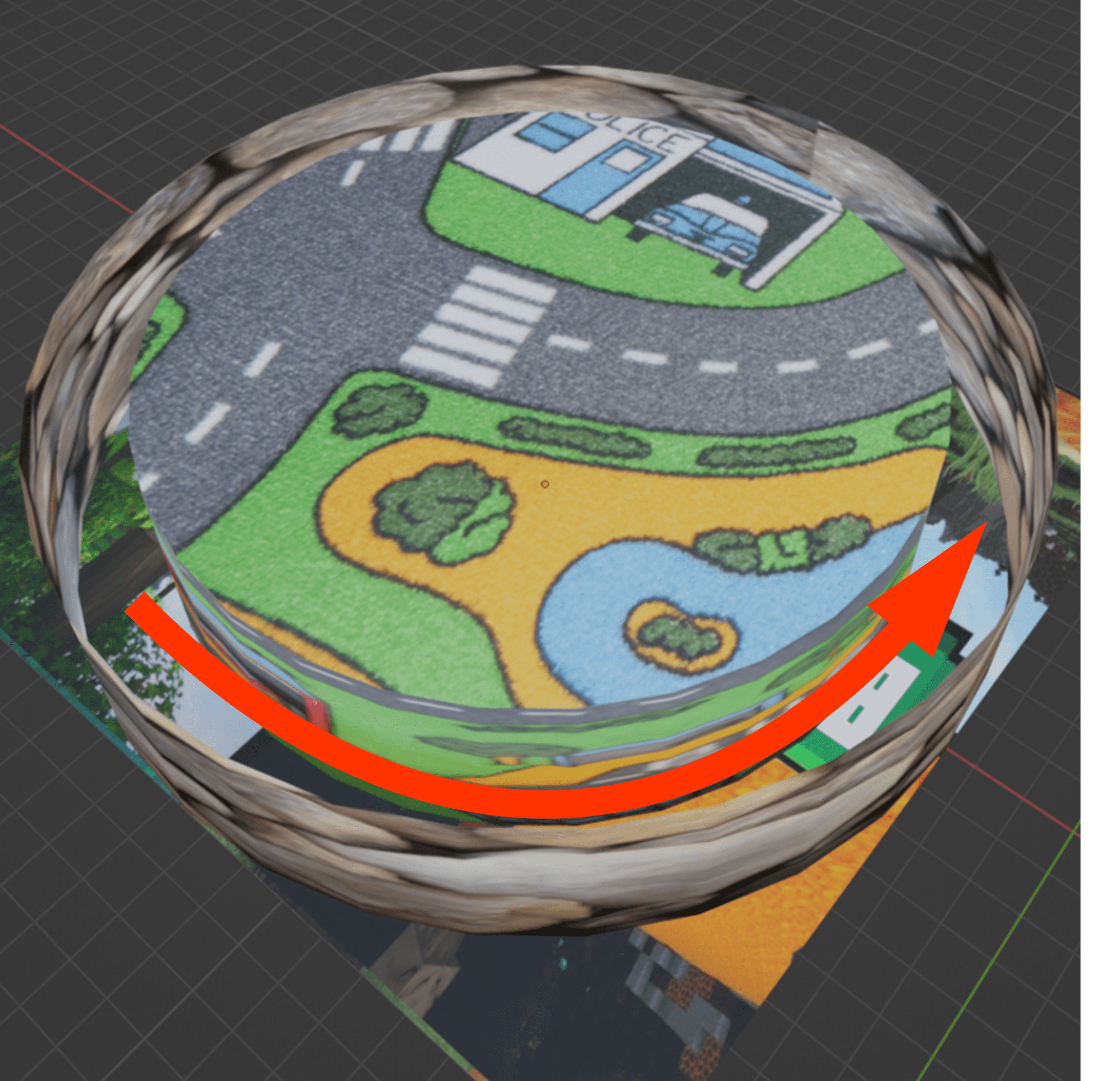}}
    \hfill
    \subfigure[Randomly distributed boxes.]{\includegraphics[width=0.47\columnwidth]{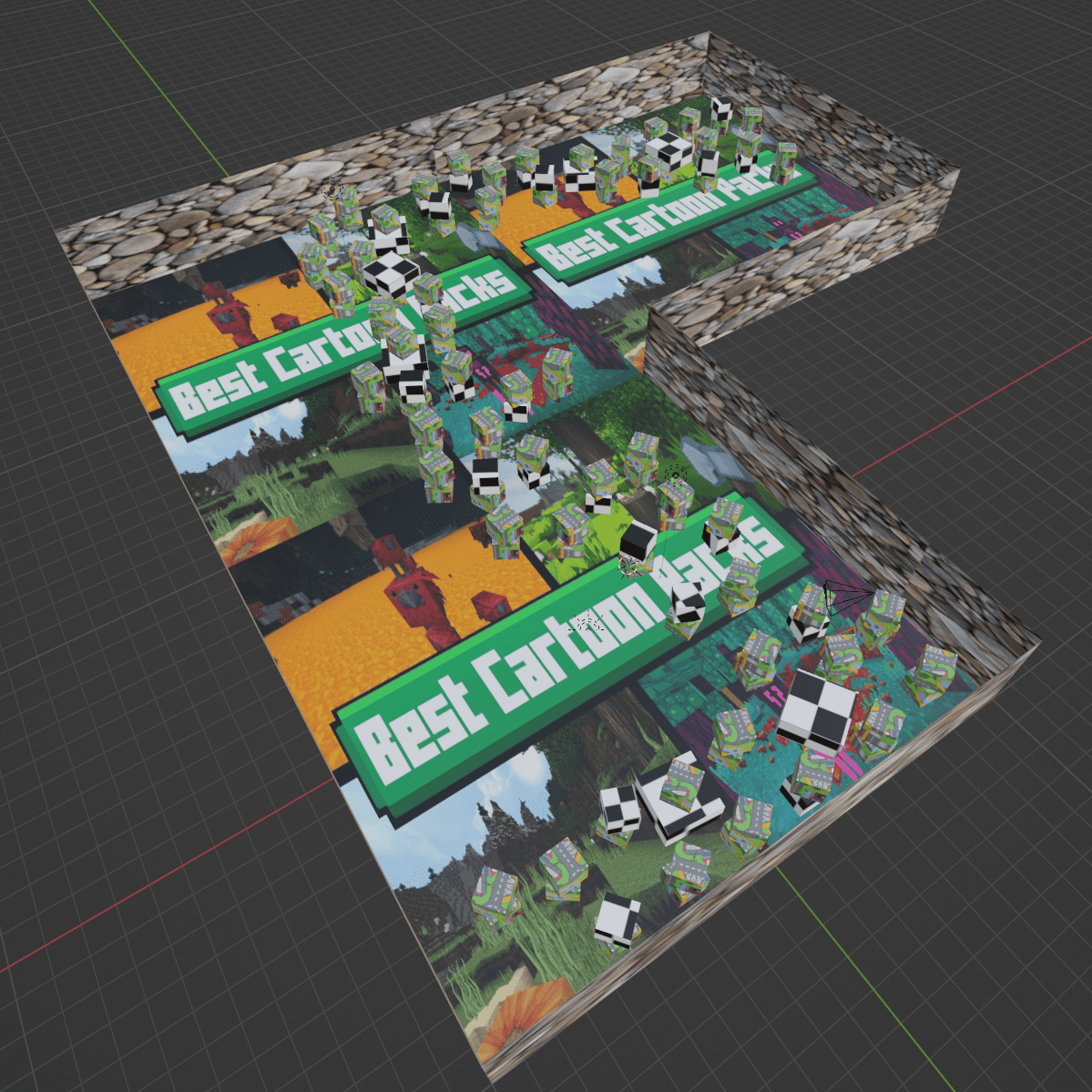}}
    \caption{Simulated scenes for synthetic data generation.
    }
    \label{fig:simulation scene}
\vspace{-0.5cm}
\end{figure}

To evaluate the proposed algorithm, we generate several sequences featuring aggressive maneuvers of a drone using the ESIM simulator~\cite{Rebecq18corl}.
Two synthetic scenes created are visualized in Fig.~\ref{fig:simulation scene}.
The first scene is a circular corridor in which the drone undergoes a high-speed detour.
The second one features a drone traveling through randomly distributed boxes.
The synthetic drone is equipped rigidly with a stereo event-based camera, a stereo standard camera, and an IMU.
Both of the event cameras and standard cameras share identical intrinsic and extrinsic parameters, simulating a pair of DAVIS~\cite{Brandli14ssc} sensors.
The spatial resolution of all the cameras is $346 \times 260$ pixel.
The standard cameras' exposure time is set to 10 ms, with a frame rate of 30 Hz.
Note that the standard cameras are only used by the comparative approaches in Sec.~\ref{subsec:evaluation result}.
Regarding the configuration of event generation model, we keep the setting as default. The eps value is $0.001$ and the contrast threshold is 0.5.
The IMU runs at 200 Hz.
The standard deviations associated to the white noise of the accelerometer and gyroscope are set to $1.86 \times 10 ^ {-2}\ m/s^2$ and $1.86 \times 10 ^ {-3} \ rad/s$, and the standard deviations of the accelerometer and gyroscope bias are set to $4.33 \times 10^{-3}\ m/s^2$ and $2.66 \times 10^{-4} \ rad/s$, respectively.

Due to the absence of publicly available datasets\footnote{Note that the aggressive data provided in \cite{Mueggler17ijrr} are captured with a monocular event camera, and thus, not applicable to our task.} that capture aggressive ego motion of a stereo event camera, we collect our own real data.
Similar to the data used in \cite{Rebecq17bmvc}, we attach a stereo event camera to a leash and spin it.
The stereo event camera is spun horizontally, vertically, and in a loopy cross shape, respectively.
An example of the spinning motion is shown at the top left of Fig.~\ref{fig: spinning leash dataset}.
A pair of DAVIS-346 cameras are used in data collection, and blur of images is clearly observed (see the top middle of Fig.~\ref{fig: spinning leash dataset}).
The real data is collected in a $9~m \times 6~m \times 3.5~m$ room equipped with a motion capture (MoCap) system. 
Some boxes are randomly distributed in the environment. 
The person spinning the stereo event camera stands in the center of the room.
By attaching several visual markers to the stereo rig, the MoCap system can report 6-DoF poses ($\bfR_{\ttW\ttM}$ and $\bft_{\ttW\ttM}$) of the marker coordinate system at 200 Hz and with sub-millimeter accuracy. 
Differentiating the position information with respect to time, we can obtain the linear velocity in the world coordinate system (\ie, a MoCap-specific reference coordinate system). 
With known transformation between the markers and the left camera, we are able to obtain the velocity in the left camera’s coordinate system by $\os{\ttB}{}\bfv{}{} = \bfR_{\ttB\ttM}\bfR_{\ttM\ttW}\os{\ttW}{}\bfv{}{}$.
Data interpolation may be required to obtain ground truth velocity at the exact moment the estimation is conducted.
The corresponding kinematic characteristics of all the data are summarized in Table.~\ref{tab:dataset characteristics}.

\begin{table}[t]
    \centering
    \caption{Characteristics of our datasets.}
    \resizebox{.47\textwidth}{!}{
        \begin{tabular}{cccc}
        \hline
        \multicolumn{1}{c}{Seq. Name} & \multicolumn{1}{c}{Scene / Motion Pattern} & \multicolumn{1}{c}{Linear Vel.} & \multicolumn{1}{c}{Angular Vel.} \\ 
        \hline
        \textit{Synthetic\_1} & Circular Corridor  & 5.68 m/s & 0.66 rad/s \\ %
        \textit{Synthetic\_2} & Circular Corridor  & 11.0 m/s & 1.45 rad/s \\ %
        \textit{Synthetic\_3} & Circular Corridor  & 18.9 m/s & 2.50 rad/s \\ %
        \textit{Synthetic\_4} & Randomly Distributed Boxes & 2.77 m/s & 2.05 rad/s \\ %
        \textit{Synthetic\_5} & Randomly Distributed Boxes & 3.19 m/s & 3.11 rad/s \\ %
        \textit{Real\_1} & Horizontal spin & 2.17 m/s & 4.82 rad/s \\ %
        \textit{Real\_2} & Vertical spin & 4.94 m/s & 13.30 rad/s \\ %
        \textit{Real\_3} & Loopy cross shape & 4.84 m/s & 11.26 rad/s \\ %
            \hline
        \end{tabular}}
    \label{tab:dataset characteristics}
\end{table}
\subsection{Evaluation on Synthetic Data}
\label{subsec:evaluation result}
To demonstrate the effectiveness of our method, we compare it against several alternative solutions that can determine the linear velocity.
The first one is the IMU integration, which only requires the initial orientation and linear velocity as a prior.
The second one is the VINS-Fusion pipeline \cite{Qin18tro}, which is widely applied in state estimation tasks of small unmanned aerial vehicles.
To make a fair comparison, the VINS-Fusion pipeline is evaluated after being successfully initialized.
The third one is a periodically re-initialized VINS-Fusion (PR-VINS) pipeline, which is to alleviate the unfairness caused by the lost camera tracking in the original VINS-Fusion.
The fourth one, called Ultimate SLAM (USLAM)~\cite{vidal2018ultimate}, is the only open-source visual-inertial odometry method for event cameras.
It uses as input the data from a standard camera, an event camera, and an IMU.
The last one (Benosman-NF) replaces the method for computing event-based normal flow in our pipeline with Benosman's method~\cite{Benosman14tnnls}.

\begin{figure}[t]
    \centering
    \subfigure[Absolute velocity estimation error of \textit{Synthetic\_3}.]{\includegraphics[width=\linewidth]{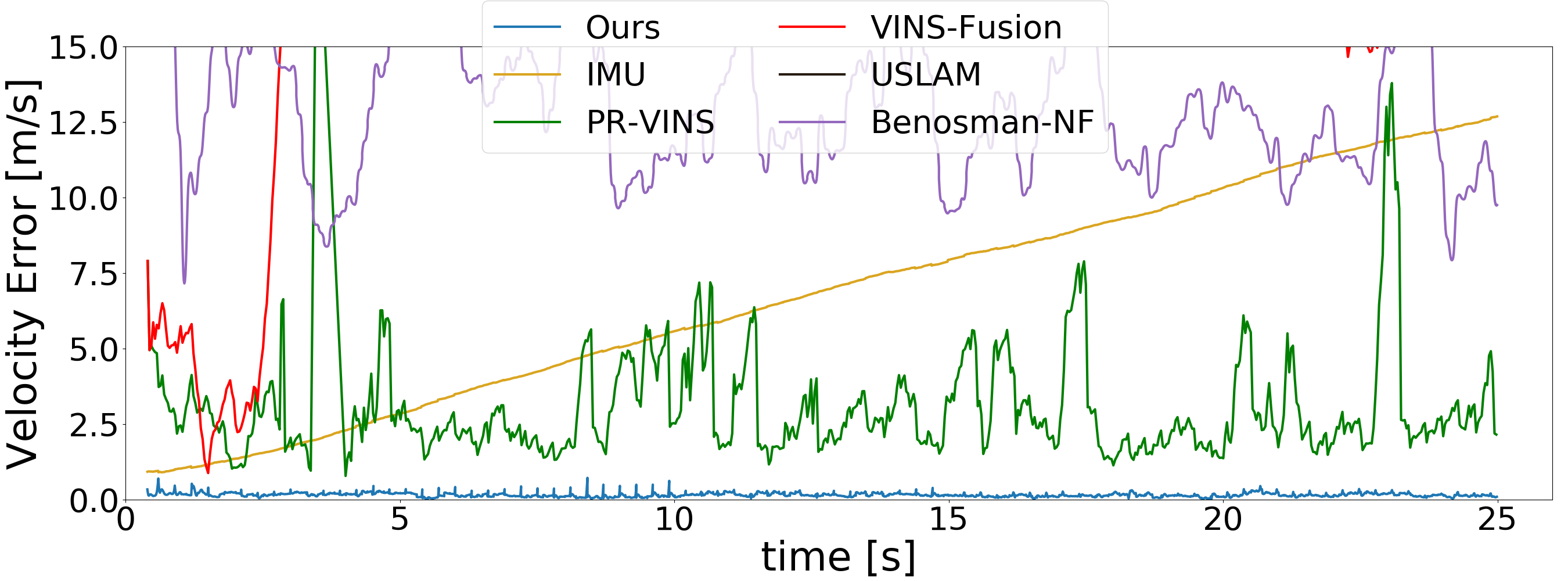}}
    \hfill
    \subfigure[Absolute velocity estimation error of \textit{Synthetic\_5}.] 
    {\includegraphics[width=\linewidth]{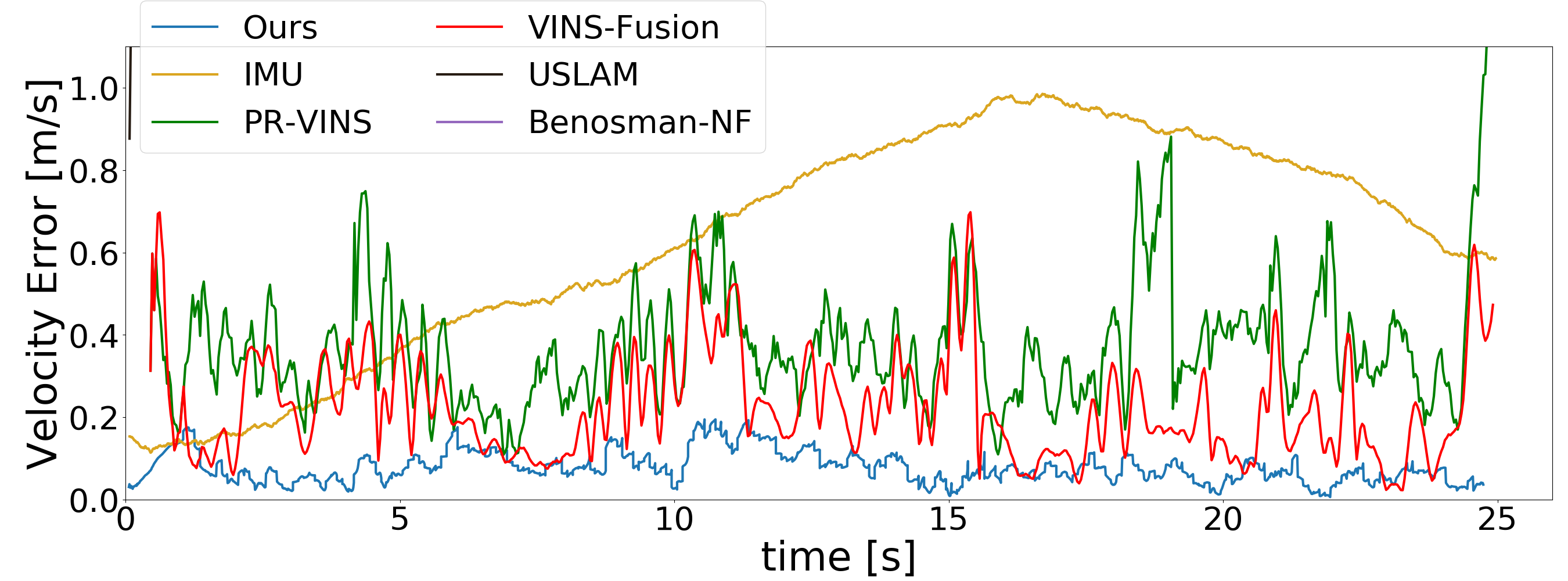}}
    \caption{\textit{Representative AVE results.}
    Note that the AVE of \textit{VINS-Fusion} and \textit{USLAM} in (a) and the AVE of \textit{Benosman-NF} and \textit{USLAM} in (b) are too large to be visualized inside the plot.}
    \label{fig:representative AVE result}
\end{figure}
\begin{table}[t]
    \Huge
    \centering
    \caption{Average AVE of each synthetic sequence (m/s).}
    \resizebox{.47\textwidth}{!}{
    \setlength{\tabcolsep}{0.25em}
        \begin{tabular}{ccccccc}
        \hline
        \multicolumn{1}{c}{Seq. Name} & IMU  & VINS-Fusion & PR-VINS & USLAM & Benosman-NF & Ours\\ 
        \hline
        \textit{Synthetic\_1} & 2.67  & 0.48  & 1.31 & Failed & 10.78 & \bf{0.09} \\ %
        \textit{Synthetic\_2} & 5.73 & 0.87  & 1.33 & 23.29 & 13.57 & \bf{0.16}\\ %
        \textit{Synthetic\_3} & 7.83  & 19.80 & 2.99 & 75.33& 13.14 & \bf{0.19} \\ %
        \textit{Synthetic\_4} & 0.54 & 0.14 & 0.27 & 19.27 & 6.71 & \bf{0.06}\\ %
        \textit{Synthetic\_5} & 0.63 & 0.23  & 0.37 & Failed & 7.75 &\bf{0.08} \\
        \hline
        \end{tabular}}
    \label{tab: ave summary}
\end{table}
To quantitatively evaluate the results, we apply as evaluation metrics the Absolute Velocity Error (AVE) and Relative Velocity Error (RVE).
The latter one is defined as the ratio of the AVE and the magnitude of ground truth velocity, namely
\begin{equation}
\text{RVE} = \frac{\vert \bfv_{gt} - \bfv_{est} \vert }{ \vert \bfv_{gt} \vert }  \times 100\%.
\label{eq:evaluate metric}
\end{equation}
To have a close-up view of each method's performance, we select as representatives the results of two sequences that feature the most aggressive maneuvers in each scene.
As illustrated in Fig.~\ref{fig:representative AVE result}, obvious accumulated errors are witnessed in the velocity estimates from the IMU-integration method.
VINS-Fusion performs normally on \textit{Synthetic\_5} but fails on \textit{Synthetic\_3} due to a failure of camera tracking caused by image blur.
The AVE of PR-VINS is typically bounded to some extent, but it is still less accurate than our method. 
The unsatisfactory results of USLAM, according to our investigation, are due to the lack of sufficient features being detected and tracked.
Note that the ego motion is notably more aggressive in the synthetic data, and thus, the USLAM can hardly achieve comparable performance as using real data (as seen in \ref{subsec: real-world experiments}), even if it has been manually initialized.
Finally, Benosman-NF typically gives bad results, which, from another perspective, justifies our derivation of event-based normal flow computation.

More detailed average AVE results can be found in Table.~\ref{tab: ave summary}.
Correspondingly, the RVE results are illustrated using box plots in Fig.~\ref{fig: rve err}.
In conclusion, our method always achieves the best performance in terms of AVE and RVE.

\begin{figure}[t]
    \centering
    \includegraphics[width=1\linewidth]{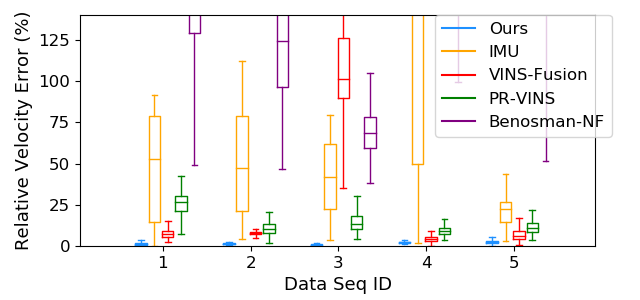}
    \caption{RVE results on our synthetic dataset.
    Note that the RVE of \textit{USLAM} is too large to be visualized inside the plot.
    }
    \label{fig: rve err}
\end{figure}
\subsection{Real-World Experiments}
\begin{table}[t]
    \Huge
    \centering
    \caption{Average AVE of each real sequence (m/s).}
    \resizebox{.47\textwidth}{!}{
    \setlength{\tabcolsep}{0.25em}
        \begin{tabular}{ccccccc}
        \hline
        \multicolumn{1}{c}{Seq. Name} & IMU  & VINS-Fusion & PR-VINS & USLAM & Benosman-NF & Ours\\ 
        \hline
        \textit{Real\_1} & 1.39 & 2.30  & 3.86 & 1.79 & 4.43 & \bf{0.72} \\ %
        \textit{Real\_2} & 9.21 & 8.43  & 7.62 & 7.48 & 4.89 & \bf{1.83}\\ %
        \textit{Real\_3} & 8.89 & 7.62 & 7.46 & 7.49 & 5.97 & \bf{1.65} \\ %
        \hline
        \end{tabular}}
    \label{tab: ave on real data}
    \vspace{-0.3cm}
\end{table}
\begin{figure}[t]
    \centering
    \includegraphics[width=1\linewidth]{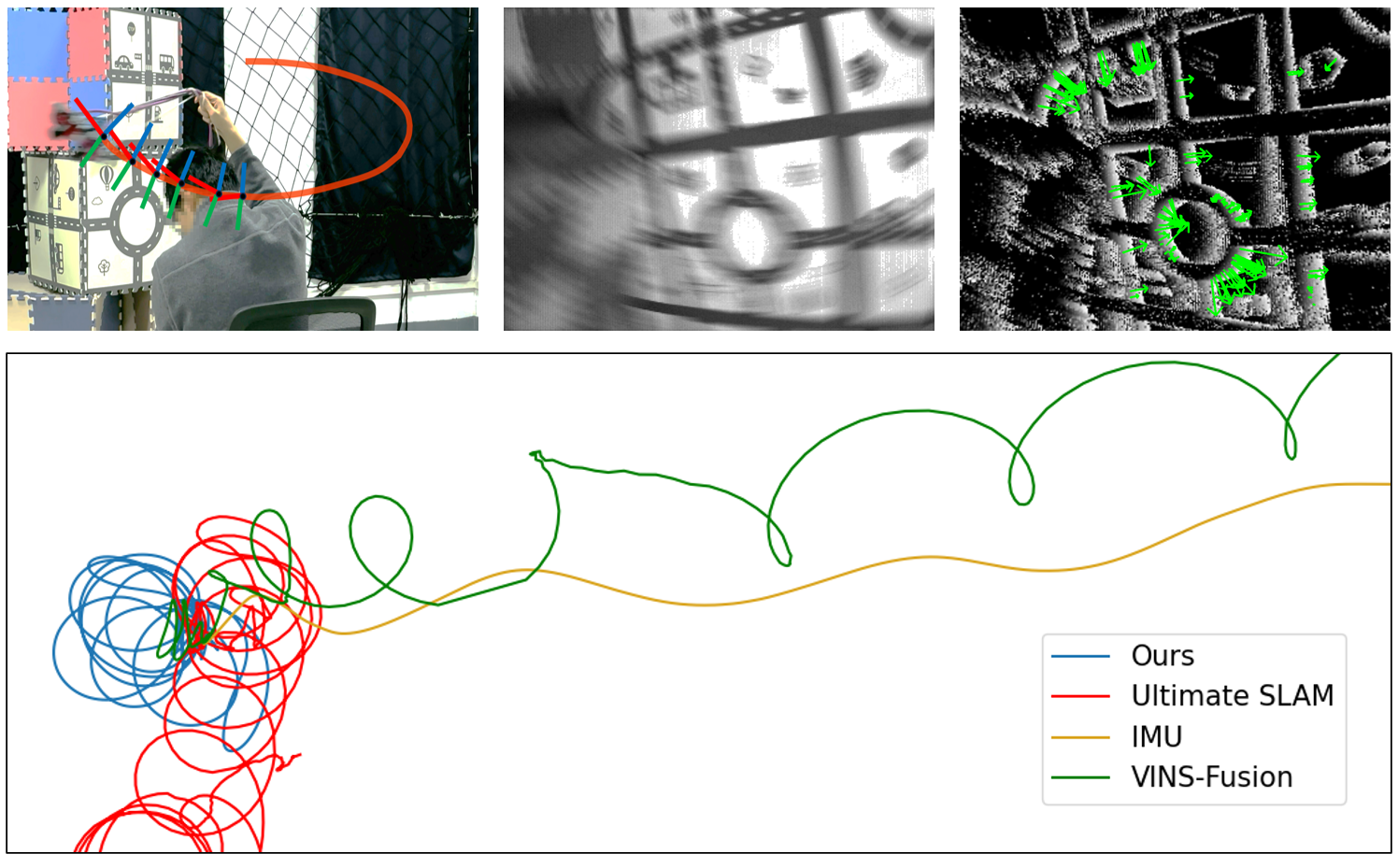}
    \caption{Self-collected data under aggressive motion.
    Top left: Person spinning horizontally a stereo event camera attached to a leash.
    Top middle: Blurry images captured by the DAVIS sensor.
    Top right: A visualization of event-based normal flows (green arrows).
    Bottom: A bird-eye view of the dead-reckoning results of different methods.}
    \label{fig: spinning leash dataset}
\end{figure}
\label{subsec: real-world experiments}

We also evaluate our method using the self-collected real data.
The quantitative results are listed in Table.~\ref{tab: ave on real data}, and the evaluation metric used is AVE.
The best results are highlighted in bold, and the conclusion is consistent with that in the evaluation on synthetic data.
Note that using the IMU alone outperforms other image-based solutions in sequence \textit{Real$\_$1}.
This is because the motion blur effect observed in sequence \textit{Real$\_$1} is much more severe compared to sequences \textit{Real$\_$2} and \textit{Real$\_$3}, leading to less accurate feature matching results on images.

Additionally, we perform a qualitative evaluation by illustrating the trajectories recovered using different methods (see Fig.~\ref{fig: spinning leash dataset}).
Both VINS-Fusion and USLAM can directly return the camera trajectories.
For the pure IMU-based solution and our method, dead reckoning is carried out.
Note that our method leads to the most reliable trajectory, USLAM performs the second best, while the others notably drift away.
The primary cause of drift in VINS-Fusion is the inaccurate feature tracking results obtained on blurry images.
The quality of feature tracking downgrades significantly under aggressive motion.
This ultimately leads to inaccurate pose estimation results that incrementally accumulate to a drift in the trajectory.
By comparison, our method does not suffer from motion blur and can exploit visual cues from event-based normal flow.
Compared to the dead-reckoning result of an IMU, our method requires only a single integration.
We demonstrate that the accurate linear velocity recovered by our method leads to better trajectories in terms of global consistency, indicating our method a better solution to the dead-reckoning task.
\subsection{Computational Performance}
\label{subsec: computational performance}

Our pipeline is implemented in C++ on ROS and all experiments are performed in real time on a desktop PC with an Intel Core 13700K CPU.
The computation time of the main modules is summarized in Table.~\ref{table: computation time}.
The normal flow and depth estimations are performed simultaneously in two independent threads.
Fed with the front-end results, the back end ultimately solves the state estimation problem using another separated thread in parallel.
The whole system can output the body-frame linear velocity up to 75 Hz.
Note that efficiency can be further improved by optimizing the code.
\begin{table}[t]
\centering
\caption{Computation time of our algorithm 
}
\begin{adjustbox}{max width=\columnwidth}
\setlength{\tabcolsep}{3pt}
\begin{tabular}{c|lccc}
\toprule
Module   & \begin{tabular}[c]{@{}c@{}} Data \\ Preprocessing\end{tabular} & \begin{tabular}[c]{@{}c@{}} Normal Flow \\ \&  Depth Est.\end{tabular}& \begin{tabular}[c]{@{}c@{}} Back-end\\ Optimization\end{tabular}  & \begin{tabular}[c]{@{}c@{}}Overall\end{tabular} \\
\midrule
Avg. Time (ms) & \multicolumn{1}{c}{3} & 10 &  10 &23\\
\bottomrule %
\end{tabular}
\end{adjustbox}
\label{table: computation time}
\end{table}

\subsection{Discussion}
\label{subsec:discussion}
\subsubsection{Limitations}
The proposed method has some limitations.
First, the event-based normal flow is computed asynchronously in the sense that the computation is triggered by the occurrence of a certain number of events.
In other words, the event-based normal flows are computed in an asynchronous batch manner.
Therefore, a batch of flow estimates may share the identical timestamps, but the time interval between two successive batches is not a constant.
Although the resulting temporal non-alignment between data can be effectively tackled by the proposed continuous-time framework, we need to investigate that if a fully asynchronous computation (i.e., an event-by-event manner) will further unlock the potential of event cameras.
The way depth information is computed will be adjusted accordingly.
Second, the uniform cubic B-spline parametrization would be replicated by a non-uniform one to deal with sudden changes of velocity.
This will be investigated as another future work.

\subsubsection{Why choose linear velocity as the only unknowns?}
The angular velocity can be definitely included in the state vector as unknowns and solved using our method.
However, it will result in worse performance in terms of accuracy and efficiency.
First, it is well known that decoupling rotation and translation estimation, if applicable, leads to better accuracy.
This is because the ambiguity in visual cues is circumvented when computing the rotation and translation separately.
Moreover, the estimated 3-D angular velocity from events, as reported in \cite{Gallego17ral}, is just as accurate as the gyroscope's measurements at its best performance.
Thus, it is reasonable to use the angular velocity measurements as input for granted.
Second, it is obvious that the involvement of angular velocity in the state vector will complicate the system, and thus, slows down the computation.
In fact, only the accurate estimates of linear velocity are key to performing a dead-reckoning operation with less drift under high-speed maneuvers.
This is why we choose linear velocity as the only unknowns.

\subsubsection{Frame-based methods vs Event-based methods}
It is also worth thinking about which is the optimal design for handling state estimation under aggressive maneuvers.
For standard frame-based cameras, the motion blur can be mitigated by adaptively reducing the integration time.
However, the adaptive exposure control complicates the system and is thus not widely applied in the community.
As for event-based cameras, they are naturally qualified for high-speed perception tasks due to the high temporal resolution.
Besides, the system delay induced by an event-based camera is even negligible compared to that caused by the blind period between successive exposures of a standard frame-based camera.
Nonetheless, we would like to clarify that although event cameras have a very high dynamic range, event-based methods are not applicable in very dark environments.

\
\section{Conclusion}
\label{sec:conclusion}

We present a novel strategy for fusing event data and inertial measurements for state estimation tasks under aggressive maneuvers.
The proposed system is called event-based visual-inertial velometer because the state estimation problem is solved on the level of the first-order kinematics, namely linear velocity.
Our front end takes as input events acquired by a pair of event cameras in stereo configuration, estimating the normal flow and corresponding depth.
The event-based normal flow directly communicates the linear velocity, closely aligned with the differential working principle of event cameras.
The back end is built on top of a continuous-time framework, which can handle data associations across heterogeneous and temporally non-aligned measurements.
Experiments demonstrate the effectiveness of our method and endorse our method as a better choice for dead-reckoning operation.
We hope our work inspires future research in event-based state estimation.
\section*{Acknowledgment}
The authors would like to thank Prof. Guillermo Gallego for the discussion at the early stage of this research,
and Dr. Yi Yu for proof reading.
We also thank the anonymous reviewers for their suggestions, which led us to improve the paper.
This work was supported by the National Key Research and Development Project of China under Grant 2023YFB4706600.

\bibliographystyle{IEEEtran}
\bibliography{main}

\end{document}